\pdfoutput=1
\documentclass{article}

\usepackage[final]{neurips_2024}

\usepackage[utf8]{inputenc} 
\usepackage[T1]{fontenc}    
\usepackage{hyperref}       
\usepackage{url}            
\usepackage{booktabs}       
\usepackage{amsfonts}       
\usepackage{nicefrac}       
\usepackage{microtype}      
\usepackage{xcolor}         
\usepackage{amsmath} 
\usepackage{comment}
\usepackage{graphicx}
\usepackage{tcolorbox} 
\usepackage{caption} 
\DeclareCaptionLabelFormat{outputlabel}{Output~#2} 
\usepackage{longtable} 

\title{Application of Contrastive Learning on ECG Data: Evaluating Performance in Japanese and Classification with Around 100 Labels}

\author{Junichiro~Takahashi \quad JingChuan Guan \quad 
\textbf{Masataka Sato} \quad
\textbf{Kaito Baba} \quad \\
\textbf{Kazuto Haruguchi} \quad 
\textbf{Daichi Nagashima} \quad 
\textbf{Satoshi Kodera} \quad 
\textbf{Norihiko Takeda} \\
  Department of Cardiovascular Medicine\\
  The University of Tokyo Hospital, Tokyo, Japan\\
  \texttt{kodera@tke.att.ne.jp} \\
}

\begin{document}
\maketitle
\begin{abstract}
The electrocardiogram (ECG) is a fundamental tool in cardiovascular diagnostics due to its powerful and non-invasive nature. One of the most critical usages is to determine whether more detailed examinations are necessary, with users ranging across various levels of expertise. Given this diversity in expertise, it is essential to assist users to avoid critical errors. Recent studies in machine learning have addressed this challenge by extracting valuable information from ECG data. Utilizing language models, these studies have implemented multimodal models aimed at classifying ECGs according to labeled terms. However, the number of classes was reduced, and it remains uncertain whether the technique is effective for languages other than English.
To move towards practical application, we utilized ECG data from regular patients visiting hospitals in Japan, maintaining a large number of Japanese labels obtained from actual ECG readings. Using a contrastive learning framework, we found that even with 98 labels for classification, our Japanese-based language model achieves accuracy comparable to previous research. This study extends the applicability of multimodal machine learning frameworks to broader clinical studies and non-English languages.
\end{abstract} 
\section{Introduction}
Electrocardiograms (ECGs) provide crucial information about the electrical activity of the heart, usually obtained from 12-lead measurement device, and play a significant role in detecting various heart diseases. 
Due to their simplicity, ECGs have been widely used as a diagnostic tool for many years~\cite{fye1994history}. They are recorded in a wide range of facilities, from clinics to general hospitals and university hospitals. 
The results of these ECGs are used by professionals with varying levels of expertise, ranging from cardiologists to non-internal medicine physicians, and even nurses.
ECG interpretation is complex because of many observation results, and the results of interpretation could vary significantly depending on the interpreter's level of expertise~\cite{kashou2023ecg}.
Therefore, the development of AI systems to assist in the interpretation of ECGs and bridge the gap in expertise is an important area of research.

There are already studies on medical multimodal AI such as LLaVA-Med~\cite{li2024llava}, which has been developed for healthcare based on language models. This model includes data such as X-ray images but does not yet support ECG which is a type of data composed by 12 time-series. Research on multimodal machine learning models that have learned from ECG data is limited to a few models such as MedGemini~\cite{saab2024capabilities}, and further investigation is needed on how to utilize ECGs in the field of machine learning. In particular, how machine learning models can process ECG data is a crucial area of study. 

CLIP~\cite{radford2021learning} has acquired knowledge about the relationships between different modalities through pretraining on a large amount of data.
Following this, there has been research that conducted pretraining using both ECG and language data~\cite{li2024frozen}, 
allowing for partially zero-shot classification about previously unseen categories~\cite{socher2013zero}. 
There are also studies that report improved performance by enhancing clinical knowledge in LLMs through reinforcement prompt engineering, utilizing a clinically validated knowledge database created by external experts~\cite{liu2024zero}.
Another study reported performance improvements by generating digital twins of ECGs using GANs~\cite{goodfellow2014generative} and extracting ECG features~\cite{hu2024personalized}. 
These studies often simplify the labels to categories such as five classes.
However, ECG interpretation in real clinical settings is complex, requiring the accurate reading of a greater number of labels from the ECG.
Additionally, the ECG dataset~\cite{liu2018open, wagner2020ptb} uses English labels, and it is unclear whether the same performance can be achieved in other languages.
Therefore, in practical applications, where a wide range of reading results is required and various languages are spoken, we need a machine learning model that can handle more comprehensive labeling and multiple languages.

Aiming for real-world implementation,
we constructed a multimodal ECG model leveraging the data obtained from patients who visit Japanese hospitals for usual medical examination.
We used enough number of Japanese labels which are utilized in normal hospital works and created by multiple cardiology specialists.
The evaluation is conducted through the classification task where partially zero-shot task is included.
\section{Method}
\begin{figure}[t]
  \centering
  \includegraphics[width=0.7\linewidth]{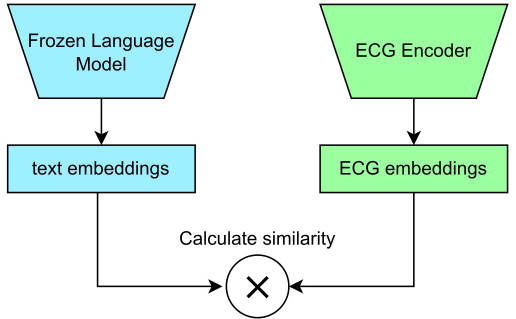}
  \caption{The overall schematics of our model. The encoder of MedLlama3-JP-v2text~\cite{MedLLama3-JP-v2} is employed as the frozen lanuguage model. ResNet1d-18~\cite{he2016deep} is adopted as the ECG encoder.  The text embeddings and ECG embeddings are denoted as $\textbf{t}_i$ and $\textbf{e}_i$, respectively.}
  \label{fig:model_schematics}
\end{figure}
\subsection{Frozen pretrained language models}
In the previous study~\cite{li2024frozen}, ClinicalBERT~\cite{alsentzer2019publicly}, pretrained on the MIMIC-III dataset~\cite{johnson2016mimic} from BioBERT~\cite{lee2020biobert}, was used as a language model with medical knowledge. In this study, we decided to select a Japanese language model based on two key criteria. The first criterion is that we should select autoregressive models. In the previous study, a BERT model~\cite{devlin2018bert} was used for contrastive learning with ECG data. In this study, considering future applications, we trained the ECG encoder using an autoregressive language model, such as GPT~\cite{radford2018improving} or LLaMA~\cite{touvron2023llama}, in order to integrate the created ECG model into a large multimodal model. We used the last layer of the hidden layers for the language embeddings. The second criterion is that the model should have medical knowledge in Japanese. Since the data used in this study consists of Japanese medical reports, it was essential to use a Japanese medical language model. The language model was selected from among Llama3~\cite{dubey2024llama}, MMed-Llama-3~\cite{qiu2024towards} OpenBioLLM~\cite{liu2018open}, MedAlpaca~\cite{han2023medalpacaopensourcecollection}, Clinical GPT~\cite{wang2023clinicalgpt}, and MedLlama3-JP-v2~\cite{MedLLama3-JP-v2}. To evaluate each LLM, cardiology specialists posed questions related to ECG in Japanese and they assessed the answers. The model judged to have the best performance was MedLlama3-JP-v2. MedLlama3-JP-v2 is a merged model consisting of Llama 3-Swallow~\cite{Okazaki:COLM2024}, OpenBioLLM, MMed-Llama-3, and Llama-3-ELYZA-JP~\cite{elyzallama2024}. It has also achieved an accuracy of 46.6\% on IgakuQA~\cite{kasai2023evaluating}, a Japanese medical QA dataset. We chose MedLlama3-JP-v2 among the 8B models available on Hugging Face due to its superior medical language knowledge in Japanese.

\subsection{ECG encoder}
We adopted ResNet1d-18~\cite{he2016deep} model based on the findings from the previous study~\cite{liu2024etp}.
They suggested that ResNet-based models~\cite{he2016deep} outperform Vision Transformer (\textbf{ViT})~\cite{dosovitskiy2020image} in both zero-shot and linear probing tasks,
and ResNet models are more effective in capturing ECG patterns.

\subsection{Multimodal contrastive learning and classification}
We will describe the method for calculating the contrastive loss. Let the batch size be 
$N$. The output from the last hidden layer of the language model is referred to as the text embedding $\textbf{t}$. The output of ResNet1d is referred to as the ECG embedding $\textbf{e}$. Both $\textbf{t}$ and $\textbf{e}$ are processed by linear layers respectively to ensure they have the same embedding dimensions. Under this condition, the contrastive loss is calculated by treating the same pair $(\textbf{t}_i,\textbf{e}_i)$~as a positive pair and the different pair $(\textbf{t}_i,\textbf{e}_j)$~as a negative pair. The similarity between the two vectors is measured using cosine similarity~($\textbf{sim}$). The cosine similarity between the two vectors is as follows:
\begin{align}
\text{sim}(\textbf{t}, \textbf{e}) = \frac{\textbf{t}^T \textbf{e}}{\|\textbf{t}\| \|\textbf{e}\|}.
\end{align}
The contrastive loss consists of two loss functions. The first loss is the ECG-to-Text contrastive loss.
\begin{align}
{l_{i}}^{(e \rightarrow t)} = -\log{\dfrac{\exp{(\text{sim}(\textbf{t}_i, \textbf{e}_i)}/\tau)}{\sum_{j=1}^{N}\exp{(\text{sim}(\textbf{t}_i, \textbf{e}_j)}/\tau)}}
\end{align}
$\tau$ is initialized to 0.07. The second is the Text-to-ECG contrastive loss.
\begin{align}
{l_{i}}^{(t \rightarrow e)} = -\log{\dfrac{\exp{(\text{sim}(\textbf{e}_i, \textbf{t}_i)}/\tau)}{\sum_{j=1}^{N}\exp{(\text{sim}(\textbf{e}_i, \textbf{t}_j)}/\tau)}}
\end{align}
Finally, the contrastive loss is calculated as the average combination of the two losses for all positive pairs within a batch.
\begin{align}
\mathcal{L} = \dfrac{1}{N}\displaystyle \sum_{i=1}^{N}\dfrac{{l_{i}}^{(e \rightarrow t)}+{l_{i}}^{(t \rightarrow e)}}{2}
\end{align}
After pre-training, we evaluated the performance of classification tasks. First, We used the prompts similar to the labels used in training. Additionally, referring to the previous research~\cite{li2024frozen}, We created a superset of the labels to evaluate zero-shot performance. The correspondence between each observation and the superset is detailed in the Appendix Table~\ref{diagnostic-comparison-table}. We do not conduct the Form test set from~\cite{li2024frozen} because there were no corresponding labels available in our data. 

\subsection{Data}
The data used in this study consists of 37285 ECG records obtained from the University of Tokyo Hospital and Mitsui Hospital. The ECGs were recorded by Fukuda Denshi (Tokyo, JAPAN) equipment. The ECG data is sequence data with a shape of 12×5000.
The experiments were conducted using 98 labels contained in the data.
The labels were selected by two cardiologists in our team out of the 157 ECG's labels specified by the equipment of Fukuda Denshi.
We formatted these specified reports in order to make training prompts as ``This ECG shows \{reports\}.'' originally in Japanese. To avoid data leakage, ECG data from the same patients were not present across any two data splits.
At the test phase evaluating the zero-shot performance, 
we used the same labels described in the prior research~\cite{li2024frozen}, that are \textbf{Supersetdiagnosis labels}, \textbf{Rhythm labels}, and \textbf{MIT-BIT labels}.
\subsection{Implementation details}
In this study, we used Hugging Face library. The learning rate was set to $1 \times 10^{-3}$, weight decay was set to $1 \times 10^{-3}$, and the global batch size was 32. We trained our model over 200 epochs. Other hyperparameters related to training were set to the default values of the Hugging Face Trainer. Training was conducted using two NVIDIA A100-SXM4-80GB GPUs.
\section{Result}
After pretraining, we firstly evaluated the performance of the classification task by using the ECG reports in the test data as the ground truth labels. We calculated accuracy for both top-1 and top-5 predictions. The results with the top 5 scores are listed in Table \ref{Results-with-the-top-5-scores}. The overall results and results of the individual labels are detailed in the Appendix Table \ref{diagnostic-result-table}. 
\begin{table}[h]
  \caption{Results with the top 5 scores (excluding results with fewer than 10 labels)}
  \label{Results-with-the-top-5-scores}
  \centering
  \begin{tabular}{llll}
    \toprule
    \multicolumn{2}{c}{Results with the top 5 scores} \\
    \cmidrule(r){1-2}
    Labels     & Top-1 Accuracy  & Top-5 Accuracy    \\
    \midrule
    Pacemaker Rhythm & 89.41\%  & 93.73\%\\
    Left Anterior Fascicular Block     & 88.00\% & 88.00\%\\
    Normal     & 78.40\%       & 90.45\%\\
    Ventricular Couplet & 77.78\% & 77.78\%\\
    Ventricular Bigeminy & 76.92\% & 84.62\%\\
    \bottomrule
  \end{tabular}
\end{table}
The results for each label suggest that our model could correctly identify normal ECGs (normal range) with high accuracy and accurately detect pacemaker rhythms (Artificial Pacemaker Rhythm).
However, it struggled to interpret the reports quantified from ECG waveforms, such as Prolonged PR Interval and Prolonged QT Interval. 
The labels related to "Short Run of Supraventricular Premature Contractions" and "Myocardial Infarction" show a significant gap between the top-1 and top-5 accuracy. For these labels, we examined the top-5 prediction results. The result is Output 2. The outputs were originally in Japanese.
\begin{tcolorbox}[colframe=black, colback=white, sharp corners, width=\textwidth]
  \textbf{label: Short Run of Supraventricular Premature Contractions}\\
  predict: This ECG shows Ventricular Premature Contractions Couplets.\\
  predict: This ECG shows Frequent Supraventricular Premature Contractions.\\
  predict: This ECG shows Supraventricular Bigeminy.\\
  predict: This ECG shows Supraventricular Premature Contractions.\\
  predict: This ECG shows Short Run of Supraventricular Premature Contractions.\\
  \\
  \textbf{label: Suspected Inferior Wall Infarction}\\
  predict: This ECG shows Suspected Inferior Wall Infarction.
\\
  predict: This ECG shows Suspected Anterior Wall Infarction.
\\
  predict: This ECG shows Suspected Lateral Wall Infarction.
\\
  predict: This ECG shows Suspected High Posterior Wall Infarction.\\
  predict: This ECG shows Suspected Acute Inferior Wall Infarction.
\end{tcolorbox}
\captionsetup{format=plain, labelformat=outputlabel,labelsep=space, justification=centering}
\captionof{figure}{The examples of the outputs of diagnosis predictions}
\vspace{\baselineskip}
\captionsetup{format=plain, labelformat=simple,labelsep=space, justification=centering}
From this output, it can be inferred that even if the top-1 prediction does not accurately identify the label, the model is still capable of detecting the ECG reports to some extent.
In the first case, based on the top 5 outputs, the model appears to have the capability to detect the events at superior ventricles.
In the second case, the model can detect myocardial infarction.
This proposes that, although the predictions is not correct, the pretrained model seems to understand some contents of the ECG reports.
\vspace{\baselineskip}

Second, referencing the prior study~\cite{li2024frozen}, we created the superset labels: \textbf{Superclass diagnosis}, \textbf{Rhythm}, and \textbf{MIT-BIH} and then evaluated the zero-shot performance. The results are in Table \ref{Superclass-diagnosis-result-table}, Table \ref{Rhythm-result-table}, Table \ref{MIT-BIH-diagnostic-result-table}.
\begin{table}[t]
\begin{minipage}[c]{0.5\hsize}
  \caption{Superclass diagnosis result}
  \label{Superclass-diagnosis-result-table}
  \centering
  \begin{tabular}{lll}
    \toprule
    \multicolumn{2}{c}{Superclass diagnosis result} \\
    \cmidrule(r){1-2}
    Labels     & Accuracy      \\
    \midrule
    all & 64.11\%  \\
    Normal ECG     & 81.53\% \\
    Conduction Disturbance     & 89.15\%    \\
    Mycardinal Infarction & 63.55\% \\
    Hypertrophy & 42.35\%\\
    ST/T change & 45.39\%\\
    \bottomrule
  \end{tabular}
\end{minipage}
\begin{minipage}[c]{0.5\hsize}
  \caption{Rhythm result}
  \label{Rhythm-result-table}
  \centering
  \begin{tabular}{lll}
    \toprule
    \multicolumn{2}{c}{Rhythm result}   \\
    \cmidrule(r){1-2}
    Labels     & Accuracy      \\
    \midrule
    all & 78.88\%  \\
    Sinus rhythm     & 95.31\% \\
    Atrial fibrillation     & 67.09\%       \\
    Sinus tachycardia & 78.05\% \\
    Sinus arrhythmia & 43.37\%\\
    Sinus bradycardia & 63.46\%\\
    \bottomrule
  \end{tabular}
\end{minipage}
\end{table}

\begin{table}[h]
  \caption{MIT-BIH diagnostic result}
  \label{MIT-BIH-diagnostic-result-table}
  \centering
  \begin{tabular}{lll}
    \toprule
    \multicolumn{2}{c}{MIT-BIH diagnostic result} \\
    \cmidrule(r){1-2}
    Labels     & Accuracy      \\
    \midrule
    all & 76.43\%  \\
    Normal beat     & 94.52\% \\
    Left bundle branch block beat     & 87.77\%       \\
    Right bundle branch block beat & 58.64\% \\
    Atrial premature beat & 41.30\%\\
    Premature ventricular contraction & 75.26\%\\
    \bottomrule
  \end{tabular}
\end{table}

Table \ref{Rhythm-result-table} and \ref{MIT-BIH-diagnostic-result-table} show that the performance of our model approaches the previous study~\cite{li2024frozen} on the Rhythm test and MIT-BIH diagnostic test set.
In the previous study~\cite{li2024frozen}, the model recorded an accuracy of 74.60\% for Rhythm test and 79.40\% for MIT-BIH test. This indicates that the contrastive learning method is generally effective for Japanese clinical reports as well. From the superclass diagnosis test in Table \ref{Superclass-diagnosis-result-table}, the model developed for this study shows shortcomings in the reports related to hypertrophy. One reason for this result is that diagnosing hypertrophy typically requires confirmation through echocardiography and while there are criteria for evaluating hypertrophy from an ECG, their sensitivity is relatively low~\cite{reichek1981left}. 
This issue implies the importance of training with echocardiographic data. 

Based on the results of this study, there was significant variability in accuracy depending on the labels. For the labels with low accuracy, further improvement in the ECG interpretation capabilities is essential. On the other hand, we consider that linking ECG with other information for training, such as echocardiographic data, is also important and we are planning to implement this approach in the future.

In addition, ablation study was conducted.
we evaluated the classification performance of Superset labels using the Swallow model~\cite{Fujii:COLM2024, Okazaki:COLM2024}, which has not trained for medical purposes.
The results are shown in Tables \ref{Normal-Swallow-Superclass-diagnosis-result-table}, \ref{Normal-Swallow-Rhythm-result-table}, and \ref{Normal-Swallow-MIT-BIH-diagnostic-result-table}. 

In Tables \ref{Normal-Swallow-Superclass-diagnosis-result-table}, \ref{Normal-Swallow-Rhythm-result-table}, and \ref{Normal-Swallow-MIT-BIH-diagnostic-result-table},
The overall accuracy is lower than the value achieved with MedLlama3-JP-v2.
This suggests that medical knowledge within the language model contributes to learning the relationship between medical texts and ECG data. For some conditions, like ST/T change and atrial premature beat, accuracy drops to around 30\%.
Swallow, used in the ablation study, lacked knowledge about these conditions and produced hallucinations.
However, for conditions like hypertrophy, the accuracy is higher than that with MedLlama3-JP-v2, which indicates the necessity for further investigation about these specific cases.

\begin{table}[t]
\begin{minipage}[c]{0.5\hsize}
  \caption{Normal Swallow Superclass diagnosis result}
  \label{Normal-Swallow-Superclass-diagnosis-result-table}
  \centering
  \begin{tabular}{lll}
    \toprule
    \multicolumn{2}{c}{Superclass diagnosis result} \\
    \cmidrule(r){1-2}
    Labels     & Accuracy      \\
    \midrule
    all & 60.49\%  \\
    Normal ECG     & 66.82\% \\
    Conduction Disturbance     & 86.28\%    \\
    Mycardinal Infarction & 68.60\% \\
    Hypertrophy & 51.70\%\\
    ST/T change & 35.66\%\\
    \bottomrule
  \end{tabular}
\end{minipage}
\begin{minipage}[c]{0.5\hsize}
  \caption{Normal Swallow Rhythm result}
  \label{Normal-Swallow-Rhythm-result-table}
  \centering
  \begin{tabular}{lll}
    \toprule
    \multicolumn{2}{c}{Rhythm result}   \\
    \cmidrule(r){1-2}
    Labels     & Accuracy      \\
    \midrule
    all & 71.47\%  \\
    Sinus rhythm     & 83.72\% \\
    Atrial fibrillation     & 60.76\%       \\
    Sinus tachycardia & 71.95\% \\
    Sinus arrhythmia & 46.99\%\\
    Sinus bradycardia & 63.46\%\\
    \bottomrule
  \end{tabular}
\end{minipage}
\end{table}
\begin{table}[h]
  \caption{Normal Swallow MIT-BIH diagnostic result}
  \label{Normal-Swallow-MIT-BIH-diagnostic-result-table}
  \centering
  \begin{tabular}{lll}
    \toprule
    \multicolumn{2}{c}{MIT-BIH diagnostic result} \\
    \cmidrule(r){1-2}
    Labels     & Accuracy      \\
    \midrule
    all & 73.76\%  \\
    Normal beat     & 89.36\% \\
    Left bundle branch block beat     & 82.01\%       \\
    Right bundle branch block beat & 59.19\% \\
    Atrial premature beat & 28.26\%\\
    Premature ventricular contraction & 78.87\%\\
    \bottomrule
  \end{tabular}
\end{table}
\section{Conclusion}
To assist physicians who read ECG data in the field of healthcare, we have built a ECG-specific
CLIP model that interprets ECG data. Incorporating contrastive learning, a multimodal model has
been constructed using ECG data and Japanese medical reports. During the training, we adopted a medical language model with frozen parameters and found that contrastive learning between ECG
and text can effectively learn the correspondence between ECG and text in Japanese, and can also
recognize detailed reports. This suggests that pretraining with ECG data and medical reports can
efficiently extract semantic ECG features across multiple languages. The machine learning model
that interprets ECG is expected to be applied in broader ways other than assisting users engaging
in the field of healthcare. For example, a representative one is wearable device which measures the
human electrical signals in daily lives. The device could be used by everyone to detect the signs and
prevent diseases. By developing the approach used in our study, we hope the result will contribute
to those downstream applications.
\section{Acknowledgements}
This work was supported by Cross-ministerial
Strategic Innovation Promotion Program (SIP) on “Integrated Health Care System”
Grant Number JPJ012425.
\bibliographystyle{plainnat} 
\bibliography{main}
\section{Appendix} 

\begin{longtable}{p{5cm}|p{2.5cm}|p{2.5cm}|p{2.5cm}}
  \caption{Mapping between ECG Labels and Zero-Shot Labels} \label{diagnostic-comparison-table} \\
  \toprule
  \textbf{Diagnosis} & \textbf{Superclass Diagnosis} & \textbf{Rhythm} & \textbf{MIT-BIH} \\
  \midrule
  \endfirsthead

  \multicolumn{4}{l}{\small\sl Continued from previous page} \\
  \toprule
  \textbf{Diagnosis} & \textbf{Superclass Diagnosis} & \textbf{Rhythm} & \textbf{MIT-BIH} \\
  \midrule
  \endhead

  \midrule
  \multicolumn{4}{r}{\small\sl Continued on next page} \\
  \endfoot

  \bottomrule
  \endlastfoot

  Sinus Tachycardia &  &  & Sinus Tachycardia \\
  Short Run of Supraventricular Premature Contractions &  & Sinus Arrhythmia & Atrial premature beat \\
  Pacemaker Rhythm & Conduction Disturbance &  & \\
  Short PR Interval &  &  & \\
  Flat T Wave &  & ST/T change & \\
  Severe Tachycardia &  &  & \\
  Borderline Q Wave & Mycardinal Infarction &  & \\
  Sinus Arrhythmia &  &  & Sinus Arrhythmia \\
  Mild ST-T Abnormality &  & ST/T change & \\
  Negative T Wave &  & ST/T change & \\
  Prolonged PR Interval & Conduction Disturbance &  & \\
  ST-T Abnormality &  & ST/T change & \\
  Suspected Inferior Wall Infarction & Mycardinal Infarction &  & \\
  Complete Right Bundle Branch Block & Conduction Disturbance &  & Right bundle branch block beat \\
  Left Ventricular Hypertrophy with Left Atrial Enlargement & Hypertrophy &  & \\
  Possible Inferior Wall Infarction & Mycardinal Infarction &  & \\
  Frequent Ventricular Premature Contractions &  &  & Premature ventricular contraction \\
  Inferior Wall Infarction & Mycardinal Infarction &  & \\
  Supraventricular Premature Contractions &  & Sinus Arrhythmia & Atrial premature beat \\
  Second-degree Atrioventricular Block (Wenckebach) & Conduction Disturbance &  & \\
  Bradycardia &  &  & \\
  First-degree Atrioventricular Block & Conduction Disturbance &  & \\
  Suspected Anteroseptal Infarction & Mycardinal Infarction &  & \\
  Severe Bradycardia & Mycardinal Infarction &  & \\
  Atrial Fibrillation &  &  & Atrial fibrillation \\
  Poor R Wave Progression & Mycardinal Infarction &  & \\
  Left Ventricular Hypertrophy & Hypertrophy &  & \\
  Incomplete Right Bundle Branch Block & Conduction Disturbance &  & Right bundle branch block beat \\
  Abnormal Q Wave & Mycardinal Infarction &  & \\
  Intraventricular Conduction Delay & Conduction Disturbance &  & \\
  Ventricular Premature Contractions &  &  & Premature ventricular contraction \\
  Suspected Left Anterior Fascicular Block & Conduction Disturbance &  & Left bundle branch block beat \\
  Anteroseptal Infarction & Mycardinal Infarction &  & \\
  Sinus Bradycardia &  &  & Sinus Bradycardia \\
  Complete Left Bundle Branch Block & Conduction Disturbance &  & Left bundle branch block beat \\
  Mild Left Ventricular Hypertrophy with Left Atrial Enlargement & Hypertrophy &  & \\
  Supraventricular Tachycardia &  & Sinus Arrhythmia & Atrial premature beat \\
  RSR' Pattern & Conduction Disturbance &  & Right bundle branch block beat \\
  Suspected Lateral Wall Infarction & Mycardinal Infarction &  & \\
  Suspected Anterior Wall Infarction & Mycardinal Infarction &  & \\
  Lateral Wall Infarction & Mycardinal Infarction &  & \\
  Tachycardia &  &  & \\
  Suspected Mild ST-T Abnormality &  & ST/T change & \\
  Left Anterior Fascicular Block & Conduction Disturbance &  & Left bundle branch block beat \\
  Atrial Flutter &  & Sinus Arrhythmia & \\
  Suspected High Posterior Wall Infarction & Mycardinal Infarction &  & \\
  Left Atrial Enlargement &  &  & \\
  Suspected Acute Inferior Wall Infarction & Mycardinal Infarction &  & \\
  Possible Lateral Wall Infarction & Mycardinal Infarction &  & \\
  Anterior Wall Infarction & Mycardinal Infarction &  & \\
  Mild Left Axis Deviation &  &  & \\
  High Voltage (Leads Corresponding to Left Ventricle) & Hypertrophy &  & \\
  Frequent Supraventricular Premature Contractions &  & Sinus Arrhythmia & Premature ventricular contraction \\
  Right Axis Deviation &  &  & \\
  Left Axis Deviation &  &  & \\
  Possible Anteroseptal Infarction & Mycardinal Infarction &  & \\
  Left Posterior Fascicular Block & Conduction Disturbance &  & Left bundle branch block beat \\
  Supraventricular Trigeminy &  & Sinus Arrhythmia & Atrial premature beat \\
  Biventricular Hypertrophy & Hypertrophy &  & \\
  Prolonged QT Interval &  &  & \\
  Mild Left Ventricular Hypertrophy & Hypertrophy &  & \\
  Acute Anterior Wall Infarction & Mycardinal Infarction &  & \\
  Low Voltage (Limb Leads) &  &  & \\
  Severe Right Axis Deviation &  &  & \\
  Ventricular Couplet &  &  & Premature ventricular contraction \\
  Subacute Anteroseptal Infarction & Mycardinal Infarction &  & \\
  Right Atrial Enlargement &  &  & \\
  Mild Right Ventricular Hypertrophy & Hypertrophy &  & \\
  Normal & Normal ECG &  & Normal Beat \\
  Clockwise Rotation &  &  & \\
  Counterclockwise Rotation &  &  & \\
  Right Ventricular Hypertrophy & Hypertrophy &  & \\
  Ventricular Rhythm &  &  & \\
  T-wave Elevation & ST/T change &  & \\
  S1, S2, S3 Pattern &  &  & \\
  Mild ST Elevation & ST/T change &  & \\
  Ventricular Bigeminy &  &  & Premature ventricular contraction \\
  Possible Anterior Wall Infarction & Mycardinal Infarction &  & \\
  Ventricular Tachycardia &  &  & \\
  Sinoatrial Block & Mycardinal Infarction &  & \\
  Indeterminate Arrhythmia &  &  & \\
  Subacute Anterior Wall Infarction & Mycardinal Infarction &  & \\
  Subacute Lateral Wall Infarction & Mycardinal Infarction &  & \\
  Subacute Inferior Wall Infarction & Mycardinal Infarction &  & \\
  Mild Right Ventricular Hypertrophy with Left Atrial Enlargement & Hypertrophy &  & \\
  Right Ventricular Hypertrophy with Right Atrial Enlargement & Hypertrophy &  & \\
  Low Voltage (Chest Leads) &  &  & \\
  Second-degree Atrioventricular Block (Mobitz) & Conduction Disturbance &  & \\
  Mild Right Ventricular Hypertrophy with Right Atrial Enlargement & Hypertrophy &  & \\
  Right Ventricular Hypertrophy with Left Atrial Enlargement & Hypertrophy &  & \\
  Suspected Acute Lateral Wall Infarction & Mycardinal Infarction &  & \\
  Ventricular Premature Contractions Couplets &  & Sinus Arrhythmia & Atrial premature beat \\
  Ventricular Trigeminy &  &  & Premature ventricular contraction \\
  Supraventricular Bigeminy &  & Sinus Arrhythmia & Premature ventricular contraction \\
  Complete Atrioventricular Block & Conduction Disturbance &  & \\
  Possible High Posterior Wall Infarction & Mycardinal Infarction &  & \\
  Acute Lateral Wall Infarction & Mycardinal Infarction &  & \\
  Suspected Acute Anterior Wall Infarction & Mycardinal Infarction &  & \\
  
\end{longtable}

\begin{longtable}{lcc}
  \caption{Top-1 and Top-5 Accuracy for Various Diagnoses} \label{diagnostic-result-table} \\
  \toprule
  \multicolumn{1}{c}{Diagnosis (Data Counts)} & \multicolumn{1}{c}{Top-1 Accuracy} & \multicolumn{1}{c}{Top-5 Accuracy} \\
  \midrule
  \endfirsthead 

  \multicolumn{3}{l}{\small\sl Continued from previous page} \\
  \toprule
  \multicolumn{1}{c}{Diagnosis (Data Counts)} & \multicolumn{1}{c}{Top-1 Accuracy} & \multicolumn{1}{c}{Top-5 Accuracy} \\
  \midrule
  \endhead 

  \midrule
  \multicolumn{3}{r}{\small\sl Continued on next page} \\
  \endfoot 

  \bottomrule
  \endlastfoot 
  all labels (7710) & 35.91\% & 44.80\% \\
  Sinus Tachycardia (82) & 73.17\% & 75.61\% \\
  Short Run of Supraventricular Premature Contractions (4) & 0.00\% & 75.00\% \\
  Pacemaker Rhythm (255) & 89.41\% & 93.73\% \\
  Short PR Interval (62) & 38.71\% & 53.23\% \\
  Flat T Wave (444) & 22.75\% & 32.66\% \\
  Severe Tachycardia (42) & 61.90\% & 61.90\% \\
  Borderline Q Wave (123) & 18.70\% & 26.83\% \\
  Sinus Arrhythmia (63) & 1.59\% & 4.76\% \\
  Mild ST-T Abnormality (278) & 16.91\% & 28.06\% \\
  Negative T Wave (258) & 24.81\% & 34.11\% \\
  Prolonged PR Interval (150) & 40.67\% & 47.33\% \\
  ST-T Abnormality (501) & 36.13\% & 44.31\% \\
  Suspected Inferior Wall Infarction (50) & 16.00\% & 42.00\% \\
  Complete Right Bundle Branch Block (278) & 58.27\% & 63.31\% \\
  Left Ventricular Hypertrophy with Left Atrial Enlargement (22) & 18.18\% & 36.36\% \\
  Possible Inferior Wall Infarction (77)& 23.38\% & 35.06\% \\
  Frequent Ventricular Premature Contractions (22) & 68.18\% & 86.36\% \\
  Inferior Wall Infarction (61) & 42.62\% & 50.82\% \\
  Supraventricular Premature Contractions (80) & 13.75\% & 22.50\% \\
  Second-degree Atrioventricular Block (Wenckebach) (1) & 0.00\% & 100.00\% \\
  Bradycardia (12) & 8.33\% & 16.67\% \\
  First-degree Atrioventricular Block (81) & 48.15\% & 58.02\% \\
  Suspected Anteroseptal Infarction (46) & 36.96\% & 58.70\% \\
  Severe Bradycardia (5) & 40.00\% & 40.00\% \\
  Atrial Fibrillation (316) & 57.59\% & 64.56\% \\
  Poor R Wave Progression (146) & 32.19\% & 39.73\% \\
  Left Ventricular Hypertrophy (296) & 13.18\% & 25.68\% \\
  Incomplete Right Bundle Branch Block (191) & 35.60\% & 40.31\% \\
  Abnormal Q Wave (51) & 3.92\% & 7.04\% \\
  Intraventricular Conduction Delay (75) & 28.00\% & 30.67\% \\
  Ventricular Premature Contractions (156) & 10.26\% & 21.79\% \\
  Suspected Left Anterior Fascicular Block (53) & 69.81\% & 86.79\% \\
  Anteroseptal Infarction (63) & 52.38\% & 53.97\% \\
  Sinus Bradycardia (52) & 57.69\% & 61.54\% \\
  Complete Left Bundle Branch Block (59) & 59.32\% & 83.05\% \\
  Mild Left Ventricular Hypertrophy with Left Atrial Enlargement (4) & 25.00\% & 25.00\% \\
  Supraventricular Tachycardia (3) & 0.00\% & 0.00\% \\
  RSR' Pattern (75) & 24.00\% & 41.33\% \\
  Suspected Lateral Wall Infarction (28)& 0.00\% & 3.57\% \\
  Suspected Anterior Wall Infarction (53)& 7.55\% & 28.30\% \\
  Lateral Wall Infarction (79)& 27.85\% & 27.85\% \\
  Tachycardia (54)& 0.00\% & 5.56\% \\
  Suspected Mild ST-T Abnormality (40)& 15.00\% & 30.00\% \\
  Left Anterior Fascicular Block (25)& 88.00\% & 88.00\% \\
  Atrial Flutter (6)& 16.67\% & 33.33\% \\
  Suspected High Posterior Wall Infarction (5)& 0.00\% & 0.00\% \\
  Left Atrial Enlargement (174)& 6.90\% & 17.82\% \\
  Suspected Acute Inferior Wall Infarction (5)& 0.00\% & 0.00\% \\
  Possible Lateral Wall Infarction (40)& 0.00\% & 0.00\% \\
  Anterior Wall Infarction (57)& 24.56\% & 24.56\% \\
  Mild Left Axis Deviation (293)& 42.66\% & 59.39\% \\
  High Voltage (Leads Corresponding to Left Ventricle) (126)& 22.22\% & 44.44\% \\
  Frequent Supraventricular Premature Contractions (3)& 33.33\% & 33.33\% \\
  Right Axis Deviation (209)& 25.36\% & 33.01\% \\
  Left Axis Deviation (147)& 18.37\% & 30.61\% \\
  Possible Anteroseptal Infarction (5)& 0.00\% & 20.00\% \\
  Left Posterior Fascicular Block (2)& 0.00\% & 0.00\% \\
  Supraventricular Trigeminy (2)& 0.00\% & 0.00\% \\
  Biventricular Hypertrophy (17)& 0.00\% & 0.00\% \\
  Prolonged QT Interval (214)& 12.62\% & 18.22\% \\
  Mild Left Ventricular Hypertrophy (55)& 21.82\% & 30.91\% \\
  Acute Anterior Wall Infarction (4)& 0.00\% & 0.00\% \\
  Low Voltage (Limb Leads) (140)& 53.57\% & 54.29\% \\
  Severe Right Axis Deviation (48)& 18.75\% & 18.75\% \\
  Ventricular Couplet (18)& 77.78\% & 77.78\% \\
  Subacute Anteroseptal Infarction (3)& 0.00\% & 0.00\% \\
  Right Atrial Enlargement (73)& 16.44\% & 26.03\% \\
  Mild Right Ventricular Hypertrophy (39)& 0.00\% & 5.13\% \\
  Normal (639)& 78.40\% & 90.45\% \\
  Clockwise Rotation (187)& 9.63\% & 11.23\% \\
  Counterclockwise Rotation (203)& 41.38\% & 48.77\% \\
  Right Ventricular Hypertrophy (10)& 0.00\% & 10.00\% \\
  Ventricular Rhythm (1)& 0.00\% & 0.00\% \\
  T-wave Elevation (20)& 45.00\% & 55.00\% \\
  S1, S2, S3 Pattern (25)& 20.00\% & 24.00\% \\
  Mild ST Elevation (21)& 9.52\% & 14.29\% \\
  Ventricular Bigeminy (13)& 76.92\% & 84.62\% \\
  Possible Anterior Wall Infarction (12)& 0.00\% & 0.00\% \\
  Ventricular Tachycardia (3)& 0.00\% & 33.33\% \\
  Sinoatrial Block (1)& 0.00\% & 0.00\% \\
  Indeterminate Arrhythmia (4)& 0.00\% & 0.00\% \\
  Subacute Anterior Wall Infarction (3)& 33.33\% & 33.33\% \\
  Subacute Lateral Wall Infarction (4)& 75.00\% & 75.00\% \\
  Subacute Inferior Wall Infarction (8)& 62.50\% & 62.50\% \\
  Mild Right Ventricular Hypertrophy with Left Atrial Enlargement (11)& 36.36\% & 45.45\% \\
  Right Ventricular Hypertrophy with Right Atrial Enlargement (2)& 100.00\% & 100.00\% \\
  Low Voltage (Chest Leads) (19)& 31.58\% & 31.58\% \\
  Second-degree Atrioventricular Block (Mobitz) (1)& 0.00\% & 0.00\% \\
  Mild Right Ventricular Hypertrophy with Right Atrial Enlargement (2)& 100.00\% & 100.00\% \\
  Right Ventricular Hypertrophy with Left Atrial Enlargement (4)& 75.00\% & 75.00\% \\
  Suspected Acute Lateral Wall Infarction & 0.00\% & 0.00\% \\
  Ventricular Premature Contractions Couplets (2)& 66.67\% & 66.67\% \\
  Ventricular Trigeminy (2)& 0.00\% & 100.00\% \\
  Supraventricular Bigeminy (2)& 0.00\% & 50.00\% \\
  Complete Atrioventricular Block (2)& 0.00\% & 0.00\% \\
  Possible High Posterior Wall Infarction (2)& 0.00\% & 0.00\% \\
  Acute Lateral Wall Infarction (1)& 0.00\% & 0.00\% \\
  Suspected Acute Anterior Wall Infarction (2)& 0.00\% & 0.00\% \\
\end{longtable}
\end{document}